\documentclass[sigconf]{acmart}
\usepackage{algorithm}
\usepackage{algorithmic}
\usepackage{multirow}
\usepackage[
type={CC},
modifier={by},
version={4.0},
]{doclicense}
\newcommand{\vpara}[1]{\vspace{0.05in}\noindent\textbf{#1 }}
\AtBeginDocument{%
  \providecommand\BibTeX{{%
    \normalfont B\kern-0.5em{\scshape i\kern-0.25em b}\kern-0.8em\TeX}}}

\setcopyright{rightsretained} 
\copyrightyear{2021} 
\acmYear{2021} 

\acmConference[KDD '21]{Proceedings of the 27th ACM SIGKDD Conference on Knowledge Discovery and Data Mining}{August 14--18, 2021}{Virtual Event, Singapore}
\acmBooktitle{Proceedings of the 27th ACM SIGKDD Conference on Knowledge Discovery and Data Mining (KDD '21), August 14--18, 2021, Virtual Event, Singapore}\acmDOI{10.1145/3447548.3467450}
\acmISBN{978-1-4503-8332-5/21/08}
\begin{document}
\fancyhead{}
\title{Adaptive Transfer Learning on Graph Neural Networks}

\author{Xueting Han}
\affiliation{%
 \country{Microsoft Research Asia}
}
\email{chrihan@microsoft.com}

\author{Zhenhuan Huang}
\authornote{This work was done when the authors were interns at Microsoft Research Asia.}
\affiliation{%
  \country{Beihang University}}
\affiliation{%
 \country{Microsoft Research Asia}}
\email{16231192@buaa.edu.cn}

\author{Bang An}
\authornotemark[1]
\affiliation{%
  \country{University of Maryland, College Park}}
\affiliation{%
  \country{Microsoft Research Asia}}
\email{bangan@umd.edu}

\author{Jing Bai}
\affiliation{%
  \country{Microsoft Research Asia}}
\email{jbai@microsoft.com}


\renewcommand{\thefootnote}{\fnsymbol{footnote}}


\begin{abstract}
Graph neural networks (GNNs) is widely used to learn a powerful representation of graph-structured data. Recent work demonstrates that transferring knowledge from self-supervised tasks to downstream tasks could further improve graph representation. However, there is an inherent gap between self-supervised tasks and downstream tasks in terms of optimization objective and training data. Conventional pre-training methods may be not effective enough on knowledge transfer since they do not make any adaptation for downstream tasks. To solve such problems, we propose a new transfer learning paradigm on GNNs which could effectively leverage self-supervised tasks as auxiliary tasks to help the target task. Our methods would adaptively select and combine different auxiliary tasks with the target task in the fine-tuning stage. We design an adaptive auxiliary loss weighting model to learn the weights of auxiliary tasks by quantifying the consistency between auxiliary tasks and the target task. In addition, we learn the weighting model through meta-learning. Our methods can be applied to various transfer learning approaches, it performs well not only in multi-task learning but also in pre-training and fine-tuning. Comprehensive experiments on multiple downstream tasks demonstrate that the proposed methods can effectively combine auxiliary tasks with the target task and significantly improve the performance compared to state-of-the-art methods. 
\end{abstract}

\ccsdesc[500]{Computing methodologies~Learning latent representations}
\ccsdesc[500]{Computing methodologies~Transfer learning}
\ccsdesc[500]{Computing methodologies~Neural networks}

\keywords{Graph Neural Networks; Transfer Learning; Multi Task Learning; GNN Pre-Training; Graph Representation Learning}

\maketitle

\section{Introduction}
Graph neural networks (GNNs)~\cite{bruna2013spectral, gcn} attract a lot of attention in representation learning for graph-structured data. Recent work demonstrates various GNN architectures~\cite{gcn,gat,graphsage,xu2018powerful} achieve state-of-the-art performances in many graph-based tasks, such as node classification~\cite{gcn}, link prediction~\cite{zhang2018link}, and graph classification~\cite{DiffPool}. These GNNs have been proven effective to learn powerful representations in a variety of scenarios, including recommendation system~\cite{DBLP:conf/kdd/YingHCEHL18} and social network~\cite{fan2019graph}. 

GNNs are often trained in an end-to-end manner with supervision which usually requires abundant labeled data, but task-specific labels can be scarce for many graph datasets. Some recent works~\cite{hu2020gpt} study GNN pre-training and employing auxiliary tasks~\cite{hwang2020self} to solve such problems. Their goal is to transfer the learned knowledge from self-supervised tasks to downstream tasks, and these methods prove making use of abundant unlabeled data could further improve graph representation. However, due to the difference of optimization objectives and data distribution between self-supervised tasks and downstream tasks, there is an inherent gap between pre-training and fine-tuning. In addition, conventional pre-training methods are not aware of which self-supervised tasks benefit more for downstream tasks and do not make any adaptation for them. This leads to knowledge transfer not effective enough for downstream tasks. There is little exploration in improving the effectiveness of transfer learning on GNNs, fully leveraging information in self-supervision, and enabling sufficient adaptation for downstream tasks.

In this work, we explore effective knowledge transfer between self-supervised tasks and downstream tasks, which is an under-investigated problem in GNNs compared to natural language processing (NLP). Optimizing the effectiveness of knowledge transfer and alleviating the gap between the pre-trained model and the target task have been the focus of recent studies in NLP~\cite{howard2018universal,chronopoulou2019embarrassingly,gururangan2020don}.~\cite{chronopoulou2019embarrassingly} addresses this by using LM as an auxiliary task to prevent catastrophic forgetting.

Inspired by these works, we propose a new generic transfer learning paradigm that adaptively selects and combines various auxiliary tasks with the target task in fine-tuning stage to achieve a better adaptation for downstream tasks. In other words, we combine the optimization objective from the target task with multiple objectives from auxiliary tasks. This preserves sufficient knowledge captured by self-supervised tasks while improving the effectiveness of transfer learning on GNNs.

\begin{figure}[h]
  \centering
  \includegraphics[width=\linewidth]{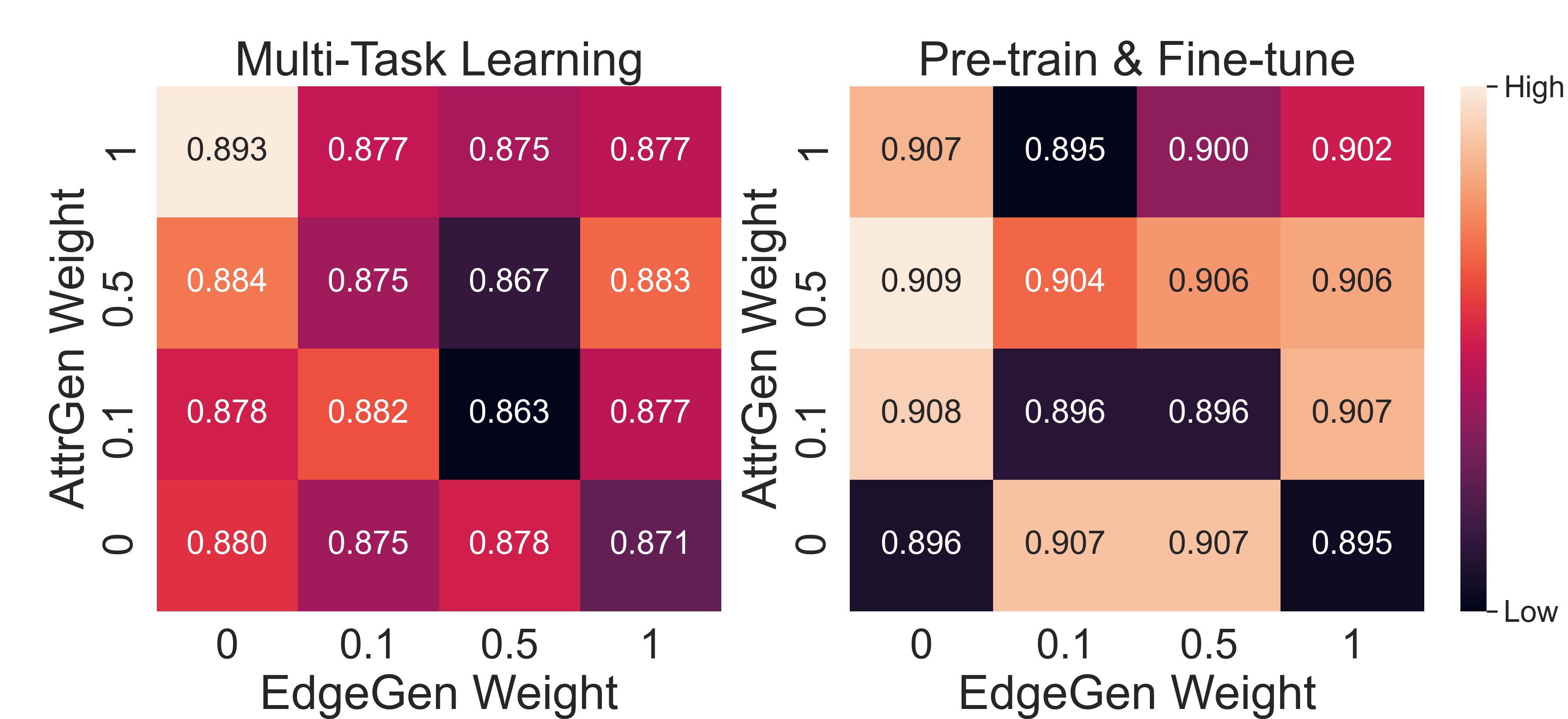}
  \caption{Performance of adding two auxiliary tasks with different weights based on HGT~\cite{hgt} (Reddit).}
  \label{fig:analysis}
\end{figure}

\textbf{Challenges.} Transferring knowledge from graph-based self-supervised tasks is a powerful tool for learning a better graph representation. Many self-supervision methods on graphs, such as DGI~\cite{velickovic2019deep}, edge generation~\cite{hu2020gpt}, attribute masking~\cite{pretrain} and meta-path prediction~\cite{hwang2020self} have been studied and their effectiveness has been proved. The basic principle, based on which additional auxiliary tasks can benefit the target task, is that auxiliary tasks share some unknown information with the target task, and to some extent, help learn features that are beneficial to the target task. However, the correlation between auxiliary tasks and the target task can be large or small and may change along with training. An auxiliary task may be helpful to the target task at the beginning of training, but may no longer be useful in the later stages of training~\cite{lin2019adaptive}. If an auxiliary task is assigned with an inappropriate weight, it may even hurt the target task, which means negative transfer~\cite{pan2009survey} occurs.

Graph preserves diverse information and downstream tasks are also diverse, single auxiliary task could not fully explore all organic signals. Previous work involving multiple self-supervised tasks in graph learning usually manually sets the weight for each task using prior intuition, e.g. assuming that all auxiliary tasks matter equally or considering the magnitude of the loss for each task~\cite{hu2020gpt}. We analyze the effect of different combinations of auxiliary tasks. Figure~\ref{fig:analysis} shows that the weights of tasks are critical, different combinations of weights lead to different results on the target task and bad combinations lead to negative transfer. We also find the same set of weights behave differently in two transfer schemes, multi-task learning from scratch and pre-training and fine-tuning (in this scheme, we add auxiliary tasks in fine-tuning stage). Therefore, the challenge of using auxiliary tasks lies in the difficulty in selecting and combining different auxiliary tasks especially when the number of tasks grows. It is still an open question that has been less explored in graph transfer learning, especially in adding auxiliary tasks in fine-tune stage after pre-train.

Some related works try to adaptively tune the weights of each task during the training process~\cite{chen2018gradnorm,liu2019end,kendall2018multi,du2018adapting,lin2019adaptive,hwang2020self}. We compared some of these methods in two schemes, one is multi-task learning and the other is adding auxiliary tasks in fine-tune stage after pre-train. We find that such weighting strategies of auxiliary tasks may be effective in multi-task learning, but may fail in pre-training and fine-tuning scheme. We argue that it's because the importance of auxiliary tasks to the target task changes in different learning stages. In multi-tasking learning, both the target task and auxiliary tasks are trained from scratch; whereas in pre-training and fine-tuning, models are already learned by abundant pre-train data.

\textbf{Contributions.} To summarize, this work makes the following major contributions.

\textbf{(i)} We propose a new generic transfer learning paradigm on GNNs which adaptively selects and combines various auxiliary tasks on graphs. It can better adapt to target tasks and improve the effectiveness of knowledge transfer. We quantify the consistency between an auxiliary task and the target task by measuring the cosine similarity between the gradients of the auxiliary task and the target task, and we design an adaptive auxiliary loss weighting model to learn the weights of auxiliary tasks by considering this signal. Besides, the weighting model could be optimized through meta-learning.

\textbf{(ii)} We conduct comprehensive experiments on both multi-task learning and pre-training and fine-tuning across multiple datasets. Our methods outperform other methods and achieve state-of-the-art results. It is worth mentioning that our methods perform well in pre-training and fine-tuning while other methods may be not effective. Our approach is not limited to GNNs scenarios but is a generic transfer learning paradigm that could be flexibly applied to various transfer learning scenarios such as in NLP or computer vision (CV). 

\textbf{(iii)} We further conduct a thorough analysis to study the advantages of our methods compared with other baseline methods. 

\section{Related Work}

\subsection{Pre-Training and Multi-Task Learning on Graph}
Pre-Training~\cite{DBLP:journals/corr/abs-1810-04805} is a common and effective way to transfer knowledge learned from related tasks to a target task to improve generalization. A general way is to pre-train a language model (LM) on a large corpus of unlabeled text and then fine-tune the model on supervised data. Recently, researchers have explored pre-training for GNNs to enable more effective learning on graphs from abundant unlabeled graph data~\cite{pretrain,hu2020gpt}. Multi-task learning~\cite{hwang2020self} is another common method to achieve shared representations by simultaneously training a joint objective of the supervised task and the self-supervised tasks.~\cite{hwang2020self} designs a group of auxiliary tasks to learn the heterogeneity of graphs and try to transfer such knowledge from auxiliary tasks to a target task. \cite{you2020does} tries to compare three schemes of self-supervision on GCNs.

However, knowledge transfer for GNN is challenging due to the differences between self-supervised tasks and downstream tasks on optimization objectives and data distribution. Optimizing the effectiveness of knowledge transfer and alleviating the gap between the pre-trained model and the target task has attracted a lot of attention in recent NLP research~\cite{howard2018universal,chronopoulou2019embarrassingly,gururangan2020don}.~\cite{howard2018universal} shows continued pre-train of an LM on the unlabeled data of a given task is beneficial for the downstream task. Towards the same direction,~\cite{chronopoulou2019embarrassingly} addresses this by using LM as an auxiliary task to prevent catastrophic forgetting. Besides,~\cite{gururangan2020don} show the importance of further adaptation with large corpora of in-domain data and task's unlabeled data. Inspired by these works, we argue that during the transfer process, knowledge from self-supervised tasks is required for both pre-training and fine-tuning.

\subsection{Self-Supervised Tasks on Graph}

Self-supervision is a promising direction to learn more transferable, generalized and robust representations. Research of self-supervised tasks on graphs attracts a lot of attention recently~\cite{kipf2016variational,graphsage,infograph,velickovic2019deep,sun2020multi}. Graph-structured data provides rich information that enables us to design self-supervised tasks from various perspectives, e.g. node attributes~\cite{pretrain,hu2020gpt}, underlying graph structure~\cite{graphsage,peng2020self,sun2020multi} or heterogeneity of graph~\cite{hwang2020self}, etc. For example, edge generation~\cite{hu2020gpt} and context prediction~\cite{pretrain} map nodes appearing in local contexts to nearby embeddings. K-hop context prediction~\cite{peng2020self} tries to preserve the global topology of graph to better characterize the similarity and differentiation between nodes. There are also methods (DGI~\cite{velickovic2019deep}, InfoGraph~\cite{infograph}, etc.) introduce Graph Informax which maximizes the mutual information between local and global representations. Besides, attribute generation~\cite{pretrain} ensure certain aspects of node attributes are encoded in the node embeddings.

Some recent work utilizes multiple self-supervised tasks together on graphs.~\cite{hu2020gpt} proposes to use attribute generation and edge generation tasks to capture the inherent dependency between node attributes and graph structure during the generative process.~\cite{hwang2020self} transfers the knowledge of heterogeneity of graph to the target task via adding auxiliary tasks of predicting meta-paths. However how to effectively select and combine different self-supervised tasks are less explored in GNNs.

Some other works~\cite{jin2020self,lu2021learning} have studied that the same auxiliary task behaves differently on different downstream tasks, and sometimes it may even cause negative-transfer. We argue that multiple kinds of information on graphs could be captured by different auxiliary tasks and their interactions and combinations provide big opportunities for us to learn a more powerful representation. 

\subsection{Auxiliary Task Weighting}

Past work has proven that, multi-task learning or auxiliary learning which simultaneously learns multiple related tasks could achieve better generalization by sharing a representation~\cite{liu2019end,hwang2020self,liu2019multi}. The goal of such methods includes achieving best performance across all tasks and only for the target task. In this paper, we only discuss the later one. The success of these approaches depends on how well aligned the auxiliary tasks are with the target task and how to combine these losses with the target task. However, the usefulness of auxiliary tasks is non-trivial to know and may change through the training.

The weight for each of the auxiliary task is usually manually set using prior intuition or through hyper-parameter tuning in above mentioned GNN self-supervised works. It becomes much harder to determine the weights as the number of auxiliary tasks grows.

Recent works study various weighting strategies to balance tasks. Some of them treat all tasks equally and adapt the weights based on the gradient norm~\cite{chen2018gradnorm} or task uncertainty~\cite{kendall2018multi}. Some other works evaluate the usefulness of auxiliary tasks to the target task and adapt the weights accordingly so that the more useful ones receive higher weights~\cite{du2018adapting,lin2019adaptive}.~\cite{du2018adapting} proposes to use gradient similarity between an auxiliary task and the target task on shared parameters to determine the usefulness of auxiliary tasks. Some recent works~\cite{maxl,hwang2020self} use meta-learning in auxiliary task selection.~\cite{maxl} proposes an elegant solution to generate new auxiliary tasks by collapsing existing classes.~\cite{hwang2020self} use meta-learning to automatically select and balance tasks and transfer the knowledge from auxiliary tasks to the target task. In this paper, we use meta-learning to learn a weigh function and the consistency of auxiliary tasks to the target task is considered in this function.

\section{Methods}

In the following, we first explain the existence of a gap between pre-training and fine-tuning, and to address the challenges, we propose our Generic Auxiliary loss Transfer Learning Framework. The goal of this framework is to effectively leverage self-supervised tasks as auxiliary tasks to help the target task. Later we introduce our Adaptive Auxiliary Loss Weighting Algorithm using Task Similarity, in which, we design an adaptive auxiliary loss weighting model to learn the weights of auxiliary tasks by quantifying task similarity and we optimize the model through meta-learning. Finally, we discuss the generality and effectiveness of our methods.

\subsection{Generic Auxiliary Loss Transfer Learning Framework}
We use $\mathcal{G} = \{ \mathcal{V}, \mathcal{E} , \mathcal{X} \}$ to denote a graph, where $\mathcal{V}$ represents the node set with $|\mathcal{V}|$ nodes, $\mathcal{E}$ stands for the edge set with $|\mathcal{E}|$ edges. Denoting $\boldsymbol{X} \in \mathbb{R}^{|\mathcal{V}| \times d}$ as the node feature matrix, and $\boldsymbol{A} \in \mathbb{R}^{|\mathcal{V}| \times |\mathcal{V}|}$ as the adjacency matrix or related variant. A GNN model $f_{\theta}$ encodes node attributes and contexts into a embedding $\boldsymbol{Z} \in \mathbb{R}^{|\mathcal{V}| \times d'}$ by iteratively aggregating messages from node's neighbors. 

A classic paradigm of pre-training on graphs is to first pre-train a general GNN model $f_{\theta}$ by self-supervised tasks on abundant unlabeled data to achieve a good initialization for downstream tasks. The network is trained with the objective as follows:

\begin{align} \label{eq:pre-train}
\theta^*, \boldsymbol{\Theta_p}^* & = \underset{\theta, \boldsymbol{\Theta_p}}{\operatorname{argmin}} \mathcal{L}_\mathrm{pre}(\theta, \boldsymbol{\Theta_p})  \notag \\
&= \underset{\theta, \boldsymbol{\Theta_p}}{\operatorname{argmin}} \mathcal{L}_\mathrm{pre}(\hat{\boldsymbol{Y}}, {h_{\boldsymbol{\Theta_p}}}({f_{\boldsymbol{\theta}}}(X, A))),
\end{align}

where $\mathcal{L}_\mathrm{pre}$ denotes the losses of pre-training tasks, $\hat{\boldsymbol{Y}}$ denotes the pseudo labels of self-supervised data and $\boldsymbol{\Theta_p}$ represents task-specific parameters for pre-train tasks, e.g. a classifier to predict pseudo labels.

Then in the fine-tuning process, shared GNN model $f_{\theta}$ is initialized by pre-trained model and fine-tuned on downstream tasks by supervised data. The objective of fine-tuning is to optimize the following:
\begin{align} \label{eq:fine-tune}
    \theta^*, \boldsymbol{\Theta_t}^* & =\underset{\theta, \boldsymbol{\Theta_t}}{\operatorname{argmin}} \mathcal{L}_\mathrm{sup}(\theta, \boldsymbol{\Theta_t})  \notag \\
    & = \underset{\theta, \boldsymbol{\Theta_t}}{\operatorname{argmin}} \mathcal{L}_\mathrm{sup}(\boldsymbol{Y}, {h_{\boldsymbol{\Theta_t}}}({f_{\boldsymbol{\theta}}}(X, A))),
\end{align}
where $\mathcal{L}_\mathrm{sup}$  is the loss of  fine-tuning on the downstream task, $\boldsymbol{Y}$ denotes labels of supervised data, and $\boldsymbol{\Theta_t}$ represents task-specific parameters for the target task. 

The goal of pre-training is to learn some general and critical features that can be transferred to downstream tasks; whereas, there is a gap between training objectives of self-supervised tasks and supervised tasks and there is not enough adaptation for downstream tasks in this paradigm of pre-training, which leaves the knowledge transfer not effective enough. As to our knowledge, there is only one work~\cite{lu2021learning} studies this problem on graphs, it proposes a method to structure the pre-training stage to simulate the fine-tuning process on downstream tasks, which directly optimizes the pre-trained model's quick adaptability to downstream tasks. We propose a new transfer learning paradigm on GNNs which would adaptively select and combine different auxiliary tasks with the target task in fine-tuning stage to achieve a better adaptation for downstream tasks and improve the effectiveness of knowledge transfer. Our method can automatically select the most useful tasks for downstream tasks from a task set and balances their weights. It's a principled and generic approach that can be applied to all kinds of auxiliary tasks.

\begin{figure}[h]
  \centering
  \includegraphics[width=\linewidth]{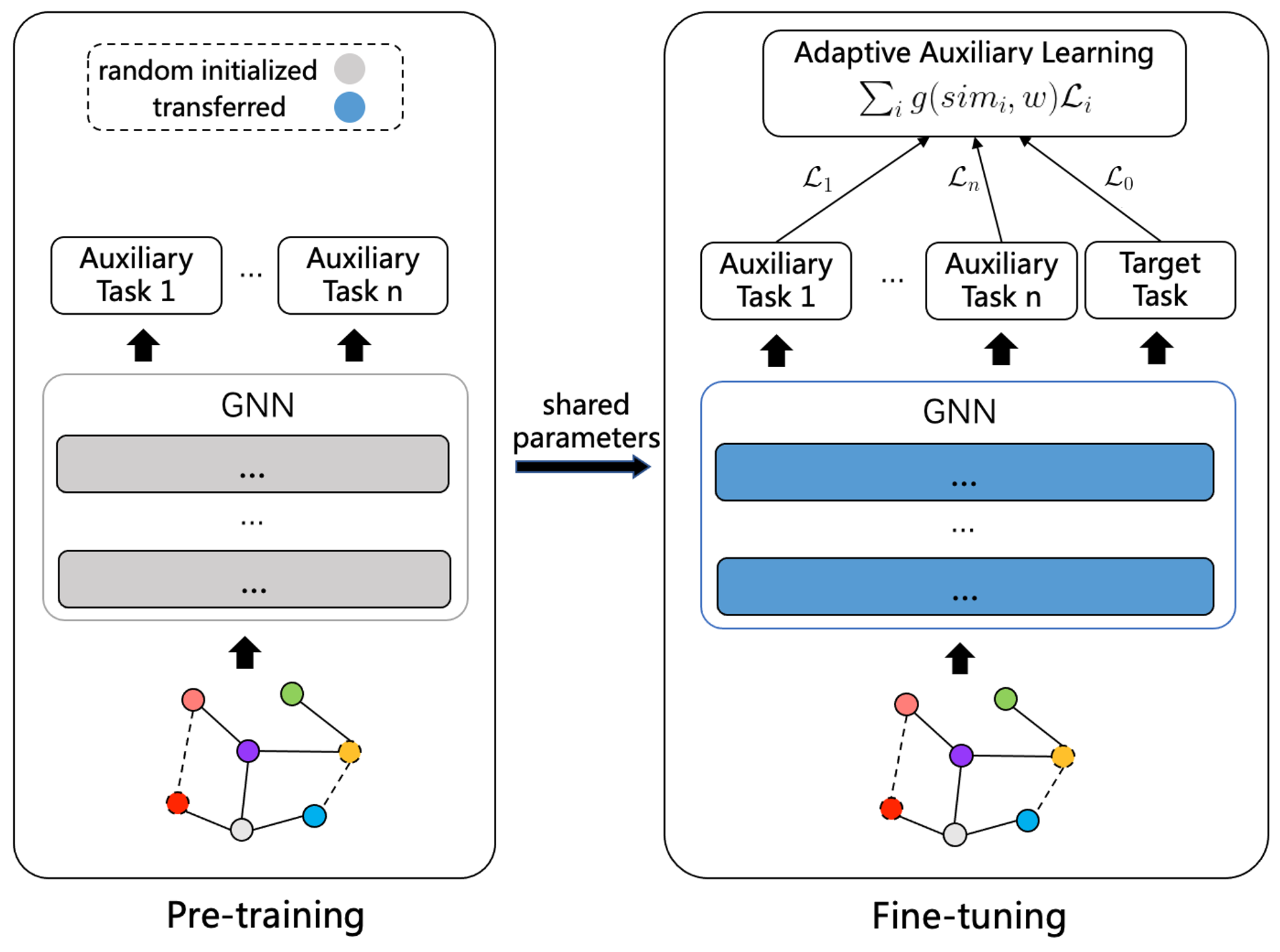}
  \caption{Generic auxiliary loss transfer learning framework.}
  \label{fig:framwork}
\end{figure}

Figure~\ref{fig:framwork} illustrates our proposed Generic Auxiliary Loss Transfer Learning Framework. In our approach, we first pre-train a GNN model by self-supervised tasks. We then transfer the parameters of the pre-trained model and fine-tune them with a joint loss. The joint loss is the weighted sum of the target task's loss and multiple auxiliary losses. We also add multiple task-specific layers for such auxiliary tasks which are randomly initialized. Adding auxiliary losses in fine-tuning can avoid catastrophic forgetting and better adapt for the target task, resulting in more powerful representations. The new joint loss for the fine-tuning process can be formulated as:
\begin{equation} \label{eq:newfinetune}
\mathcal{L}(\theta_{t})=w_{0}\mathcal{L}_{sup}(\theta_{t})+\sum_{i=1}^{K} w_{i} \mathcal{L}_{aux, i}(\theta_{t}),
\end{equation}
where $\mathcal{L}_{sup}$ is the loss of the target task that we want to complete and $\mathcal{L}_{a u x, i}$ represents the loss of auxiliary task $\boldsymbol{i}$ and $\boldsymbol{i} \in \{1, 2, ...,K\}$. The losses of both the target task and auxiliary tasks are calculated on shared model parameters $\theta_{t}$ and their task-specific layers (omitted in the above formula) at each training step $\boldsymbol{t}$. $w_{i}$ is the weight for task $\boldsymbol{i}$ which is generated by the weighting model, especially $w_{0}$ means the weight for the target task. We will introduce more details about the weighting model in Section~\ref{AUX-TS}. Assuming that we use gradient descent to update the model, the optimization of shared parameters $\theta_{t}$ on this joint loss is as follows:
\begin{equation}
\theta_{t+1}=\theta_{t}-\alpha \nabla_{\theta_{t}} \mathcal{L}(\theta_{t}).
\end{equation}
If a large number of auxiliary tasks are used, some auxiliary tasks may play a more important role than others in the feature representation learning for the target task; therefore, the weight $w_{i}$ of each auxiliary task needs to be selected carefully.

\subsection{Adaptive Auxiliary Loss Weighting using Task Similarity} \label{AUX-TS}

In previous work, the process of selecting auxiliary tasks on graphs seems to be too heuristic, and there is not any patterns to follow, which may raise some issues. First, computation will be much expensive when the number of auxiliary tasks K increases. Besides, the usefulness of each auxiliary task may change during the training; using a fixed value as weight may limit or even hurt the performance.

We propose Adaptive Auxiliary Loss Weighting using Task Similarity (AUX-TS), which quantifies the similarity of each auxiliary task to the target task and uses the information obtained after applying the auxiliary task to learn weights during training.

(1) \textbf{Weighting model quantifies the consistency of each auxiliary task to the target task} 

Our approach designs a weighting model to learn weights for different auxiliary tasks that benefit the target task most. The weighting model is a 2-layer MLP network, which takes signals that related to current data and task as input and outputs the weight.

\cite{du2018adapting} propose to use the cosine similarity of gradients between tasks to represent the relatedness between tasks and prove the effectiveness of the method. However, this method only uses the signal of cosine similarity in a rule-based way to determine whether adopt the gradients from an auxiliary task. The weights in this method cannot be optimized. Inspired by this work~\cite{du2018adapting}, our method follows the principle that a useful auxiliary task should provide gradient directions which helps to decrease the loss of the target task.

Specifically, we quantify the consistency of an auxiliary task with the target task by measuring the cosine similarity between the gradients of the auxiliary task and the target task. And we design an adaptive auxiliary loss weighting model to learn the weights of auxiliary tasks by considering this signal. We use $g([sim, type\_emb, loss]; w)$ to denote the weighting model ($g(sim; w)$ is used in the following formulas as a simplified expression), where $[sim, type\_emb, loss]$ represents all the signals that are taken as the input to weighting model, it's the concatenation of the gradient similarity, loss value and type embedding of the current task. $w$ represents the parameters of the weighting model. Since the target task is of our interest, the weighting model is optimized by minimizing the loss of the target task $\mathcal{L}_{sup}$ at each training step $t$.

(2) \textbf{Optimize weighting model via Meta-learning}

Some recent works~\cite{maxl,hwang2020self} propose the idea of using meta-learning to learn meta-parameters, which themselves are used in another learning procedure. In our method, we treat the weighting model as meta-parameters and optimize it according to the target task during the training. Specifically, we train two networks in two stages. The first stage is to optimize the weighting model which generates weights for all tasks and the second stage is to optimize the shared GNN networks on the combination of the target and auxiliary tasks in a multi-task learning scheme.

In the first stage, the weighting model is optimized by the loss of the target task, encouraging to generate a set of weights, so that if the joint loss weighted by these weights is used to train GNN, the performance of the target task could be maximized.

Gradient similarity between tasks is a strong indicator of the consistency of tasks. We denote the weighting model as $g(sim_i;w)$ which tasks gradients similarity between an auxiliary task and the target task as input:
\begin{equation}
    {sim}_{i} = \cos (\nabla_{\theta} \mathcal{L}_{sup}({\theta}), \nabla_{\theta} \mathcal{L}_{aux, i}({\theta})),
\end{equation}
where $\nabla_{\theta} \mathcal{L}_{sup}({\theta})$ is the gradient vector evaluated at $\theta$ by the loss of the target task and $\nabla_{\theta} \mathcal{L}_{aux, i}({\theta})$ is the gradient vector evaluated by the loss of auxiliary task $i$. Using $g(sim_i;w)$ to generate weight for each task, the joint loss can be represented as:
\begin{equation}
    \mathcal{L}(\theta; {w})=g(sim_0;w)\mathcal{L}_{sup}(\theta)+\sum_{i=1}^{K} g(sim_i;w) \mathcal{L}_{aux, i}(\theta)
    \label{eq:loss}.
\end{equation}
After one update of GNN by above total loss, we leverage the loss of the target task on the updated GNN to optimize the weighting model. Therefore, the weighting model is updated as follows:
\begin{equation}
 \underset{w}{\operatorname{argmin}} \mathcal{L}_{sup}(\hat{\theta}({w})).
\end{equation}
$\hat{\theta}({w})$ denotes the updated parameters of the GNN after one gradient update using the joint loss $\mathcal{L}(\theta; {w})$ (defined in Equation~\ref{eq:loss} )
\begin{equation}
    \hat{\theta}_t({w}_t) = \theta_t - \alpha \nabla_{\theta} \mathcal{L}(\theta_t; {w}_t),
\end{equation}
where $\alpha$ is the learning rate for GNN. We do not really update the parameters $\theta$, but plug it in ${L}_{sup}$ to optimize $w$.

In the second stage, GNN is trained by joint losses of both the target and auxiliary tasks, given the weights of different tasks are determined by the weighting model, we define the multi-task objective of GNN as follows:
\begin{equation}
\underset{\theta}{\operatorname{argmin}} g(sim_0;w)\mathcal{L}_{sup}(\theta)+\sum_{i=1}^{K} g(sim_i;w) \mathcal{L}_{aux, i}(\theta).
\end{equation}
We train both networks in an iterative manner until convergence. Assuming that we use gradient descent to update both networks, our proposed method is illustrated in Algorithm~\ref{alg:atl}.
\begin{algorithm}[htb]
\caption{AUX-TS}
\label{alg:atl}
\begin{algorithmic}[1]
\renewcommand{\algorithmicrequire}{\textbf{Input:}}
\REQUIRE
training data for the target task/auxiliary tasks $D^{sup}, D^{aux}$;\\
learning rate: $\alpha, \beta$, max iterations $N$;\\
the number of auxiliary tasks $K$;
\newline
\STATE \textbf{Initialize} ${\theta}_0$, ${w}_0$, ${sim}_{0} = 1$
\FOR {$t=0$ to $N$}
    \STATE fetch one batch of training data: $D^{sup}$, $D^{aux}$
    \STATE \# step 1: optimize the weighting model via meta-learning
    \STATE  $D^{sup(train)}, D^{sup(meta)} \leftarrow \text{Split}(D^{sup})$ 
    \FOR {$i=1$ to $K$} 
        \STATE\# compute gradient similarity
        \STATE ${sim}_{i} \leftarrow \cos (\nabla_{\theta_t} \mathcal{L}_{sup}({\theta}_{t}), \nabla_{\theta_t} \mathcal{L}_{aux, i}({\theta}_{t}))$
    \ENDFOR
    \STATE $\mathcal{L}(\theta_t;{w}_t) \leftarrow g(sim_0;w_t)\mathcal{L}_{sup}(\theta_{t}) + $\\
    $\qquad \qquad \qquad \sum_{i=1}^{K}g(sim_i;w_t)\mathcal{L}_{aux, i}(\theta_{t})$
    \STATE compute:  
    $\hat{\theta}_t({w}_t) \leftarrow \theta_t - \alpha \nabla_{\theta} \mathcal{L}(\theta_t; {w}_t)$ with $D^{sup(train)} \cup D^{aux}$
    \STATE Update ${w}_{t+1} \leftarrow {w}_{t} - \beta \nabla_{{w}}\mathcal{L}_{sup}(\hat{\theta}_t({w}_t))$ with $D^{sup(meta)}$ 
    \STATE \# step 2: optimize the GNN via joint loss
    \FOR {$i=1$ to $K$} 
        \STATE\# compute gradient similarity
        \STATE ${sim}_{i} \leftarrow \cos (\nabla_{\theta_t} \mathcal{L}_{sup}({\theta}_{t}), \nabla_{\theta_t} \mathcal{L}_{aux, i}({\theta}_{t}))$
    \ENDFOR
    \STATE $\mathcal{L}(\theta_{t}) \leftarrow g(sim_0;w_{t+1})\mathcal{L}_{sup}(\theta_{t}) + $\\
    $\qquad \qquad \quad \sum_{i=1}^{K} g(sim_i;w_{t+1}) \mathcal{L}_{aux, i}(\theta_{t})$
    \STATE Update $\theta_{t+1} \leftarrow \theta_{t}-\alpha \nabla_{\theta_{t}} \mathcal{L}(\theta_{t})$ with $D^{sup} \cup D^{aux}$ \\
\ENDFOR
\end{algorithmic}
\end{algorithm}
\subsection{Discussion}
First, our proposed Generic Auxiliary Loss Transfer Learning Framework could alleviate the gap between pre-training and fine-tuning and improve the effectiveness and adaptability of transferring knowledge to downstream tasks significantly. In addition, it has good generality and can be applied to all kinds of self-supervised tasks set and all kinds of GNN models.

Secondly, our proposed AUX-TS algorithm dynamically adjusts the weights for each auxiliary task during the training via meta-learning. We verify our algorithm in two transfer learning schemes. One is multi-task learning, in which GNN model is initialized randomly. The other is pre-training and fine-tuning, in which GNN model is initialized by relatively well pre-trained parameters. Considering the difference between these two learning schemes, the usefulness of auxiliary tasks in different learning stages also changes. In multi-task learning, some auxiliary tasks may contribute more to the joint loss. After the knowledge of the pre-trained model is transferred, the target task may become more important and some of the auxiliary tasks may become less important along with training. Our method uses gradient similarity between an auxiliary task and the target task as an informative signal to learn the usefulness of an auxiliary task. So that it could adaptively generate weights and perform well in both schemes.

\section{Evaluation}
We evaluate our proposed methods on four datasets to prove the effectiveness of our Generic Auxiliary loss Transfer Learning  Framework and adaptive auxiliary task weighting algorithm (AUX-TS). To evaluate the generality of our methods, we conduct experiments on two auxiliary tasks, multiple GNN models, and two transfer learning schemes. At last, we conduct a comprehensive analysis to study the advantages of our methods compared with other baseline methods.
\subsection{Datasets and Baseline}

{\bf Datasets\quad} We utilize four public datasets from different domains to evaluate the effectiveness of our methods. The detailed statistics are summarized in Table~\ref{tab:data}.
\begin{itemize}
\item \textbf{Open Academic Graph (OAG)}~\cite{tang2008arnetminer,DBLP:conf/kdd/ZhangLTDYZGWSLW19} is the largest public academic dataset to date which contains more than 178 million nodes and 2.236 billion edges. Each paper in OAG is associated with fields (e.g. Computer Science, Medicine) and publication date ranges from 1900 to 2019. We use Paper-Field (L2) prediction on the Computer Science dataset (CS) as the downstream task and evaluate the performance by MRR and NDCG, where MRR is the default one.
			
\item \textbf{Reddit}~\cite{graphsage} is a social network constructed with Reddit posts in different topical communities. We use the node classification task as the downstream task, which is to classify Reddit posts to different communities and use micro F1-score as the evaluation metric.
			
\item \textbf{Book-Crossing}~\cite{wang2018ripplenet,wang2019knowledge} contains book ratings from 0 to 10 in the Book-Crossing community. The link prediction of \textit{user-item} is used as the downstream task and is measured by AUC.
		
\item \textbf{Last-FM}~\cite{wang2018ripplenet,wang2019knowledge} contains musician listening information from a set of 2 thousand users from Last-FM online music system. The link prediction task is used as the downstream task and is measured by AUC.
\end{itemize}

\begin{table}[ht]
\begin{center}
\caption{Datasets statistics.}
\label{tab:data}
\resizebox{\linewidth}{!}{
\begin{tabular}{clrrrr}
\toprule
 & Datasets & \# Nodes & \# Edges &  \\
  \midrule
 \multirow{2}{*}{Node classification} & OAG\_CS    & 1,116,163  & 28,427,508    \\
& Reddit  & 232,965  & 114,615,892   \\
 \midrule
\multirow{2}{*}{Link prediction} & Last-FM    & 15,084  & 73,382  \\
& Book-Crossing & 110,739 & 442,746  \\

\bottomrule
\end{tabular}}
\end{center}
\end{table}

{\bf Baseline\quad} We compare the following approaches:
\begin{itemize}
\item \textbf{No Auxiliary Losses} denotes our base learning algorithm, which updates the model by the target task only instead of any auxiliary tasks. 
			
\item \textbf{Fixed Weights} updates the model by both the target task and auxiliary tasks, with a group of pre-defined fixed weights. We assign all the tasks the same weight of 1.

\item \textbf{SELAR} learns weight functions that consider the loss value to select auxiliary tasks and balance them with the target task via meta-learning.
		
\item \textbf{Cosine Similarity (unweighted)} uses gradient similarity to determine whether adopt the gradients from an auxiliary task \cite{du2018adapting}. Specifically, it combines the gradients from an auxiliary task and the target task to update the shared parameters $\theta$ only if $\cos (\nabla_{\theta} \mathcal{L}_{sup}({\theta}), \nabla_{\theta} \mathcal{L}_{aux}({\theta})) \geq 0$.

\item \textbf{AUX-TS,} as described in Algorithm 1. We have two variants, AUX-TS (one sim) corresponds to adding current task's similarity in weighting model, and AUX-TS (all sims) corresponds to adding all tasks' similarities in weighting model.
\end{itemize}

Our proposed auxiliary weighting methods can be applied to two transfer learning schemes, one is multi-task learning (MTL), the other is pre-training and fine-tuning (P\&F). We evaluate our methods with baseline methods on these two schemes. In all the experiments, we report the mean of test performance based on three independent runs.

\begin{table*}[]
\centering
\caption{Experimental results of different methods over OAG\_CS and Reddit datasets.}
\label{tab:oag}
\centering
\begin{tabular}{l|cccc|cc}
\toprule
Dataset &  \multicolumn{4}{c}{OAG\_CS} & \multicolumn{2}{|c}{Reddit} \\ \midrule
&\multicolumn{2}{c}{MTL} & \multicolumn{2}{c}{P\&F} & \multicolumn{1}{|c}{MTL} & P\&F \\
Metric &  NDCG & MRR & NDCG & MRR & F1 score & F1 score \\ \midrule
No Auxiliary Losses &  0.4387$\pm$0.0054 & 0.4941$\pm$0.0091 & 0.4584$\pm$0.0039 & 0.5284$\pm$0.0065 &  0.8474$\pm$0.0072 & 0.8605$\pm$0.0073 \\
Fixed Weights &  0.4470$\pm$0.0040 & 0.5117$\pm$0.0077 & 0.4594$\pm$0.0033 & 0.5319$\pm$0.0079 & 0.8497$\pm$0.0056 & 0.8655$\pm$0.0050 \\
Cosine Similarity &  0.4419$\pm$0.0028 & 0.5031$\pm$0.0095 & 0.4584$\pm$0.0037 & 0.5330$\pm$0.0089 & 0.8469$\pm$0.0082 & 0.8648$\pm$0.0064 \\
SELAR & 0.4512$\pm$0.0041 & 0.5197$\pm$0.0073 & 0.4617$\pm$0.0037 & 0.5382$\pm$0.0063 & 0.8499$\pm$0.0111 & 0.8723$\pm$0.0099 \\ \midrule
AUX-TS (all sims) &  0.4518$\pm$.0033 & 0.5215$\pm$0.0076 & \textbf{0.4634$\pm$0.0033} & \textbf{0.5412$\pm$0.0074} &  {0.8591$\pm$0.0116} & \textbf{0.8744$\pm$0.0092} \\
AUX-TS (one sim) &  \textbf{0.4524$\pm$0.0033} & \textbf{0.5218$\pm$0.0073} & \textbf{0.4634$\pm$0.0033} & 0.5406$\pm$0.0073 &  \textbf{0.8595$\pm$0.0092} & {0.8738$\pm$0.0090} \\ \bottomrule
\end{tabular}
\end{table*}


\begin{table*}[]
\centering
\caption{Experimental results of different methods over Last-FM and Book-Crossing datasets.}
\label{tab:book}
\begin{tabular}{ccc|cccc|cc}
\toprule
Dataset & Task & Base GNNs & \multicolumn{1}{l}{\begin{tabular}[c]{@{}l@{}}No Auxiliary\\ Losses\end{tabular}} & \multicolumn{1}{l}{\begin{tabular}[c]{@{}l@{}}Fixed \\ Weights\end{tabular}} & \multicolumn{1}{l}{\begin{tabular}[c]{@{}l@{}}Cosine \\ Similarity\end{tabular}} & \multicolumn{1}{l|}{SELAR} & \multicolumn{1}{l}{\begin{tabular}[c]{@{}l@{}}AUX-TS\\ (all sims)\end{tabular}} & \multicolumn{1}{l}{\begin{tabular}[c]{@{}l@{}}AUX-TS\\ (one sim)\end{tabular}} \\ \midrule
\multirow{8}{*}{Last-FM} & \multirow{4}{*}{MTL} & GCN & 0.7762 & 0.7985 & 0.7770 & 0.7999 & 0.8003 & \textbf{0.8015} \\
 &  & GAT & 0.7983 & 0.8132 & 0.7990 & 0.8214 & 0.8226 & \textbf{0.8289} \\
 &  & GIN & 0.7969 & 0.8125 & 0.7855 & 0.8258 & \textbf{0.8265} & 0.8252 \\ \cmidrule{3-9} 
 &  & Avg. Gain & - & 2.23\% & -0.42\% & 3.19\% & 3.29\% & \textbf{3.55\%} \\ \cmidrule{2-9} 
 & \multirow{4}{*}{P\&F} & GCN & 0.8448 & 0.8426 & 0.8393 & \textbf{0.8461} & 0.8440 & 0.8430 \\
 &  & GAT & 0.8395 & 0.8441 & 0.8411 & 0.8493 & \textbf{0.8501} & 0.8497 \\
 &  & GIN & 0.8283 & 0.8374 & 0.8275 & 0.8470 & 0.8515 & \textbf{0.8528} \\ \cmidrule{3-9} 
 &  & Avg. Gain & - & 0.46\% & -0.19\% & 1.19\% & \textbf{1.31\%} & \textbf{1.31\%} \\ \midrule
\multirow{8}{*}{Book-Crossing} & \multirow{4}{*}{MTL} & GCN & 0.6816 & 0.6850 & 0.6683 & 0.6976 & 0.7050 & \textbf{0.7169} \\
 &  & GAT & 0.6597 & 0.6362 & 0.6531 & 0.6493 & 0.6307 & \textbf{0.6774} \\
 &  & GIN & 0.6724 & 0.7338 & 0.7030 & \textbf{0.7501} & 0.7437 & 0.7453 \\ \cmidrule{3-9} 
 &  & Avg. Gain & - & 2.05\% & 0.53\% & 4.14\% & 3.26\% &  \textbf{6.25\%}\\ \cmidrule{2-9} 
 & \multirow{4}{*}{P\&F} & GCN & 0.6818 & 0.6906 & 0.6811 & \textbf{0.7165} & 0.7098 & 0.7131 \\
 &  & GAT & 0.6716 & 0.6746 & 0.6674 & 0.6825 & 0.6818 & \textbf{0.6839} \\
 &  & GIN & 0.7044 & 0.7471 & 0.7169 & 0.7429 & 0.7469 & \textbf{0.7495} \\ \cmidrule{3-9} 
 &  & Avg. Gain & - & 2.65\% & 0.37\% & 4.09\% & 3.92\% & \textbf{4.31\%} \\ \bottomrule
\end{tabular}
\end{table*}

\subsection{Experiments Setup and Results}
\vpara{Transfer learning with edge generation and attribute generation.} GCN~\cite{gcn} is chosen as the base model for OAG\_CS and Reddit. We use edge generation and attribute generation as auxiliary tasks and follow the hyper-parameter settings in GPT-GNN~\cite{hu2020gpt}, and the full labeled data is used in our experiments. For multi-task learning, we train the model from scratch for 500/500 epochs, while for pre-training and fine-tuning, we first pre-train a GNN model for 20/50 epochs and then fine-tune it on downstream tasks for 500/500 epochs on OAG\_CS/Reddit. We adopt the Time Transfer settings~\cite{hu2020gpt} for the pre-training and fine-tuning on OAG\_CS, specifically, we use data from different time spans for pre-training (before 2014) and fine-tuning (since 2014). We randomly split Reddit, 70\% of which is used for pre-training, and the other 30\% is used for fine-tuning (Train/Valid/Test Data each account for 10\%). Table~\ref{tab:oag} summarizes the performance of different methods on OAG\_CS and Reddit.
			
\vpara{Transfer learning with meta-path prediction}. We evaluate our methods on Book-Crossing and Last-FM with multiple GNN architectures: GCN~\cite{gcn}, GAT~\cite{gat} and GIN~\cite{xu2018powerful}. We use five auxiliary tasks of meta-path prediction and follow most hyper-parameter settings in SELAR~\cite{hwang2020self}, hyper-parameters such as learning rate and batch size are tuned using validation sets for all models. For multi-task learning, we train the model from scratch for 100/50 epochs, while for pre-training and fine-tuning, we first pre-train a GNN model for 100/50 epochs and then fine-tune it on downstream tasks for 100/50 epochs on Last-FM/Book-Crossing. Since the sizes of these two datasets are relatively small, we pre-train and fine-tune on the same graph. Results on Book-Crossing and Last-FM are summarized in Table~\ref{tab:book}. Details of implementation and auxiliary tasks can be found in Appendix A/B.

Overall, the proposed methods significantly enhance the performance for all downstream tasks on all datasets. Compared with No Auxiliary Losses baseline, our new transfer learning framework which adaptively selects and combines various auxiliary tasks with the target task effectively improves target tasks. Especially under the pre-training and fine-tuning, it alleviates the gap between pre-training and fine-tuning and achieves a better adaptation for target tasks. On average, our methods achieve (relative) performance gains of 5.61\%/2.42\% and 1.43\%/1.61\% over No Auxiliary Losses baseline under MTL/P\&F respectively on OAG\_CS and Reddit, and 3.55\%/1.31\% and 6.25\%/4.31\% on Last-FM and Book-Crossing. These numbers prove the effectiveness of our framework. Moreover, AUX-TS consistently outperforms other auxiliary task weighting methods on all datasets.

Comprehensive experiments evaluate the generality of our methods.

\vpara{Different transfer schemes:} considering the difference between multi-task learning and pre-training and fine-tuning, we verify our methods on both transfer schemes. As observed from Table~\ref{tab:oag}, Fixed Weights and Cosine Similarity are not as effective in pre-training and fine-tuning as in multi-task learning on OAG, while AUX-TS consistently perform well under both schemes.

\vpara{Diff auxiliary tasks sets:} we evaluate on two sets of auxiliary tasks and find AUX-TS greatly improves the results, which proves that AUX-TS can be flexibly applied to all kinds of combinations of tasks.

\vpara{Different GNN model:} Table~\ref{tab:book} shows that AUX-TS consistently improves the performance of target tasks for all three GNNs compared with other baseline methods.

\begin{figure}[h]
  \centering
  \includegraphics[width=\linewidth]{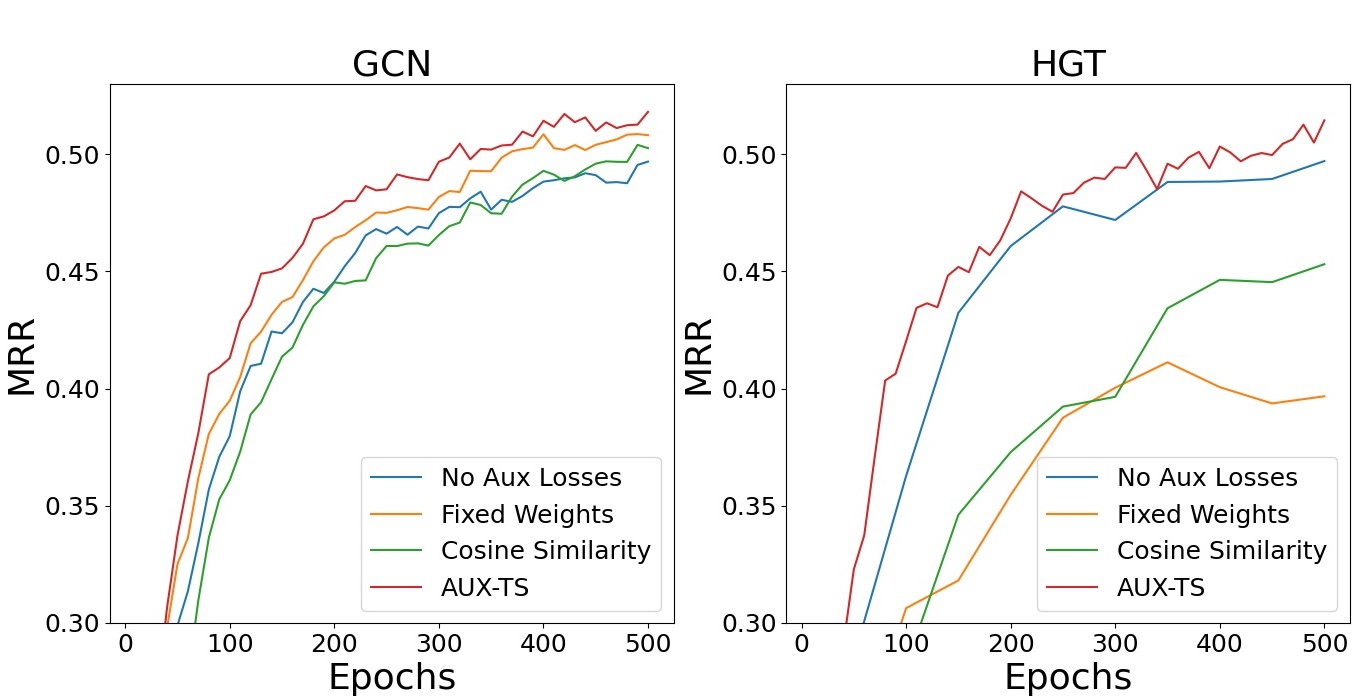}
  \caption{Performance curves of different methods on GCN and HGT under MTL (OAG\_CS).}
  \label{fig:perf}
\end{figure}
\begin{figure}[h]
  \centering
  \includegraphics[width=\linewidth]{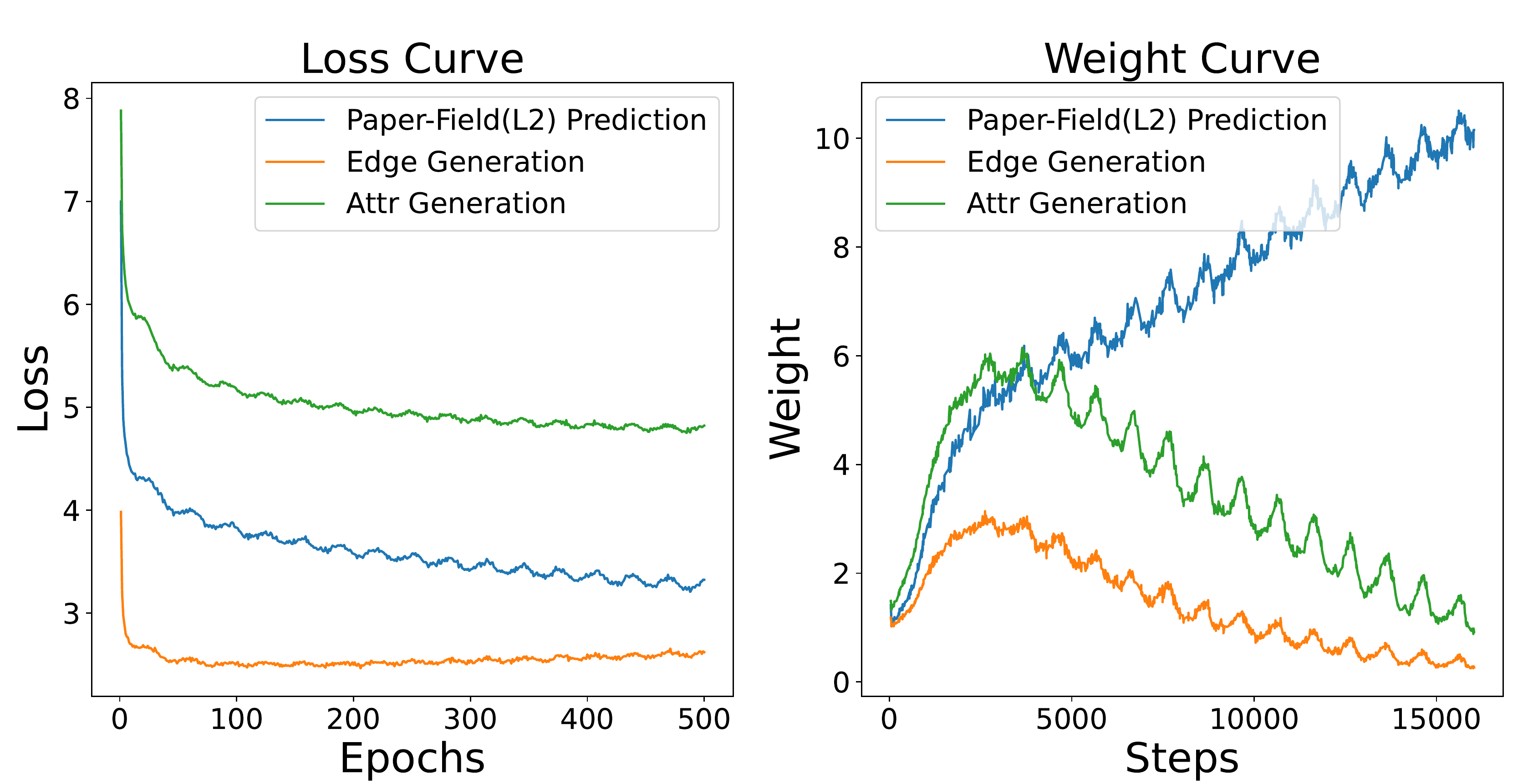}
  \caption{Loss and weight curves of AUX-TS for different tasks under P\&F (OAG\_CS).}
  \label{fig:loss}
\end{figure}

\begin{figure}[h]
  \centering
  \includegraphics[width=\linewidth]{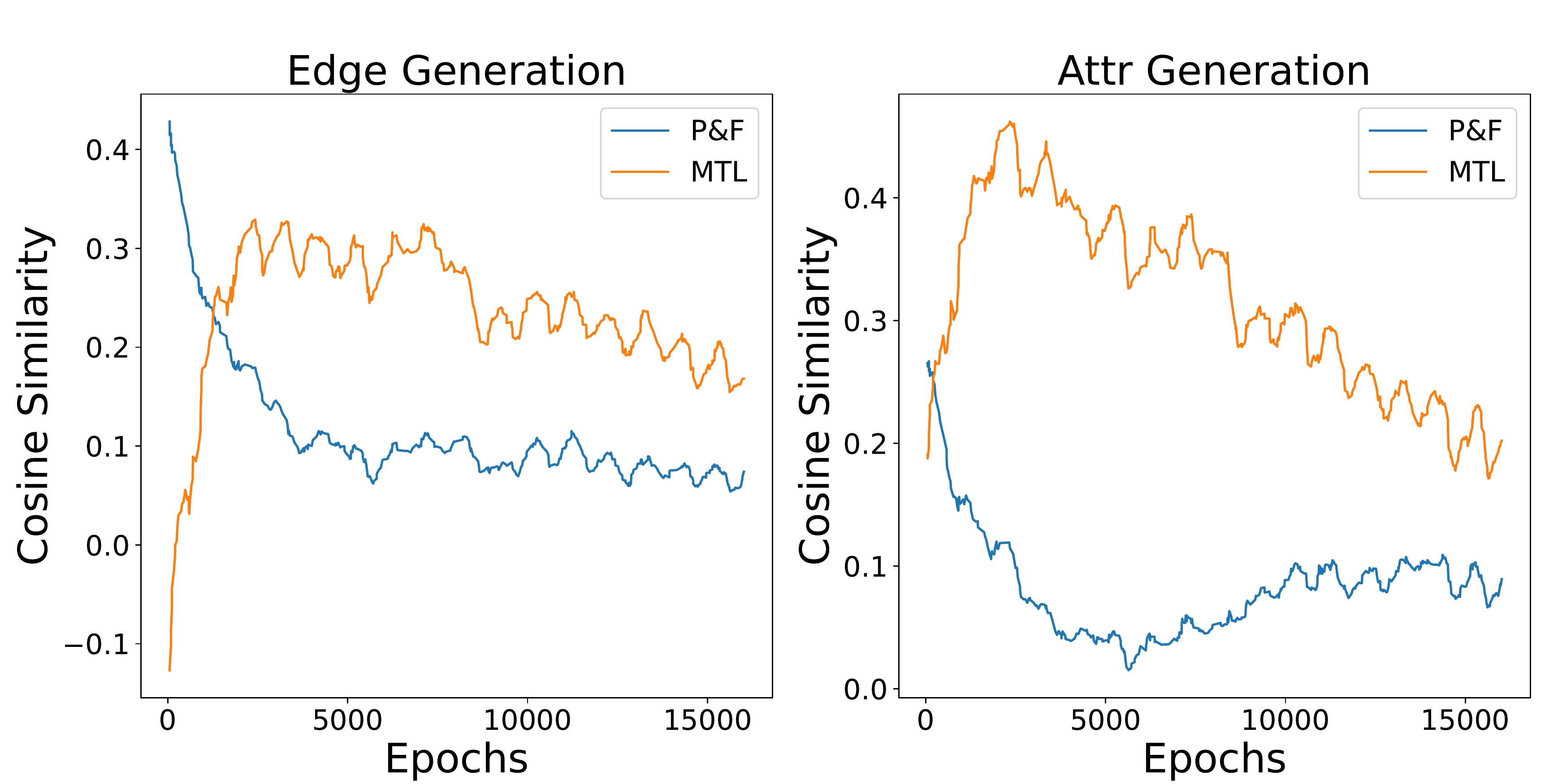}
  \caption{Gradient similarity curves of Fixed Weights under MTL and P\&F (OAG\_CS).}
  \label{fig:cos}
\end{figure}

\subsection{Further Analysis}
The effectiveness and generality of our approach are proved above. We conduct further analysis to understand why our methods work and to study the advantages of our methods compared to other baseline methods. The following analysis is based on our best-performed experiments.

\textbf{AUX-TS avoids negative transfer.} We compare the performance (MRR) curve of our methods with other baseline methods on GCN~\cite{gcn} and HGT~\cite{hgt} in Figure~\ref{fig:perf}. 
HGT is a heterogeneous model and is much more complex than GCN. AUX-TS performs better than Fixed Weights and Cosine Similarity on both models, and these two methods hurt the performance on HGT. We can observe that 1) auxiliary tasks are not always helpful to the target task; 2) negative transfer is more likely to occur on the more complex model; and 3) AUX-TS can successfully avoid negative transfer in both models.

\textbf{AUX-TS adaptively assigns appropriate weights for different tasks during the training.} From the loss and weight curves of AUX-TS under pre-training and fine-tuning in Figure~\ref{fig:loss}, we observe that the weights of different tasks change during the training. The weight of the target task increases with the training, while the weights of the two auxiliary tasks increase first and then decrease and AUX-TS assigns higher weights to the attribute generation task. Meanwhile, both the losses of the target task and the attribute generation task decrease continually while the loss of the edge generation task starts to increase at a certain stage instead. 
Although the loss of the edge generation task increases, we have achieved the best results in the target task. This proves that AUX-TS could learn the usefulness of different auxiliary tasks and can adaptively select auxiliary tasks with appropriate weights for the target task.

\textbf{Gradient similarity is a good indicator to describe the consistency between tasks, and it changes with learning.} The curve of gradient similarity for Fixed Weights under two transfer schemes (as shown in Figure~\ref{fig:cos}) reflects that, in multi-task learning, the gradient similarities between auxiliary tasks and the target task rise rapidly in the early training stage, and then decrease with the progress of training. In fine-tuning stage, after the knowledge of the pre-trained model is transferred to the downstream task, the gradient similarities drop to a much smaller value. Fixed weights or inappropriate weights of auxiliary tasks may not be helpful or even cause negative transfer, especially in fine-tuning stage. This further explains why some other weighting methods are not effective on OAG\_CS when the model is pre-trained on a huge graph. Whereas AUX-TS can perform well in both schemes since it considers the consistency of an auxiliary task to the target task.

Cosine Similarity method also uses the gradient similarity to adjust the weight. However, the gradient similarity fluctuates greatly, and there is not a simple positive correlation between the usefulness of auxiliary tasks and gradient similarity. Not only data with large gradient similarity can help the target task, but small ones can also be beneficial. Cosine similarity uses only those whose gradient similarity is greater than zero. So this rule-based method is not effective enough and far from optimal to utilize this important signal to learn weights.
Whereas in AUX-TS it's learnable via meta-learning and the advantage of AUX-TS is that it can dynamically update the weighting model to adapt to the latest training situation. We find that in multi-task learning, the gradient similarity is relatively large (-0.3 $\sim$ 0.9), the weight change curve and the gradient similarity curve show a certain positive correlation, but in pre-training and fine-tuning, the similarity is relatively small (-0.2 $\sim$ 0.4), this correlation becomes weak or even the opposite.

\begin{table}[h]
\caption{Performance (NDCG/MRR) of using different signals in weighting model on four groups of loss scales (OAG\_CS, P\&F).}
\label{tab:ablation}
\centering
\resizebox{\linewidth}{!}{
\begin{tabular}{l|ccc}
\toprule
Loss Scales & No Auxiliary Losses & SELAR & AUX-TS (all sims) \\ \midrule
edge=1 attr=1 & \multirow{4}{*}{0.4584/0.5284} & 0.4617/0.5382 & \textbf{0.4634/0.5412} \\
edge=1 attr=5 & & 0.4586/0.5332 & \textbf{0.4637/0.5426}\\
edge=5 attr=1 & & 0.4545/0.5258 & \textbf{0.4641/0.5415}\\
edge=5 attr=5 & & 0.4569/0.5287 & \textbf{0.4642/0.5431}\\ \bottomrule
\end{tabular}
}
\end{table}

\begin{figure}[h]
  \centering
  \includegraphics[width=\linewidth]{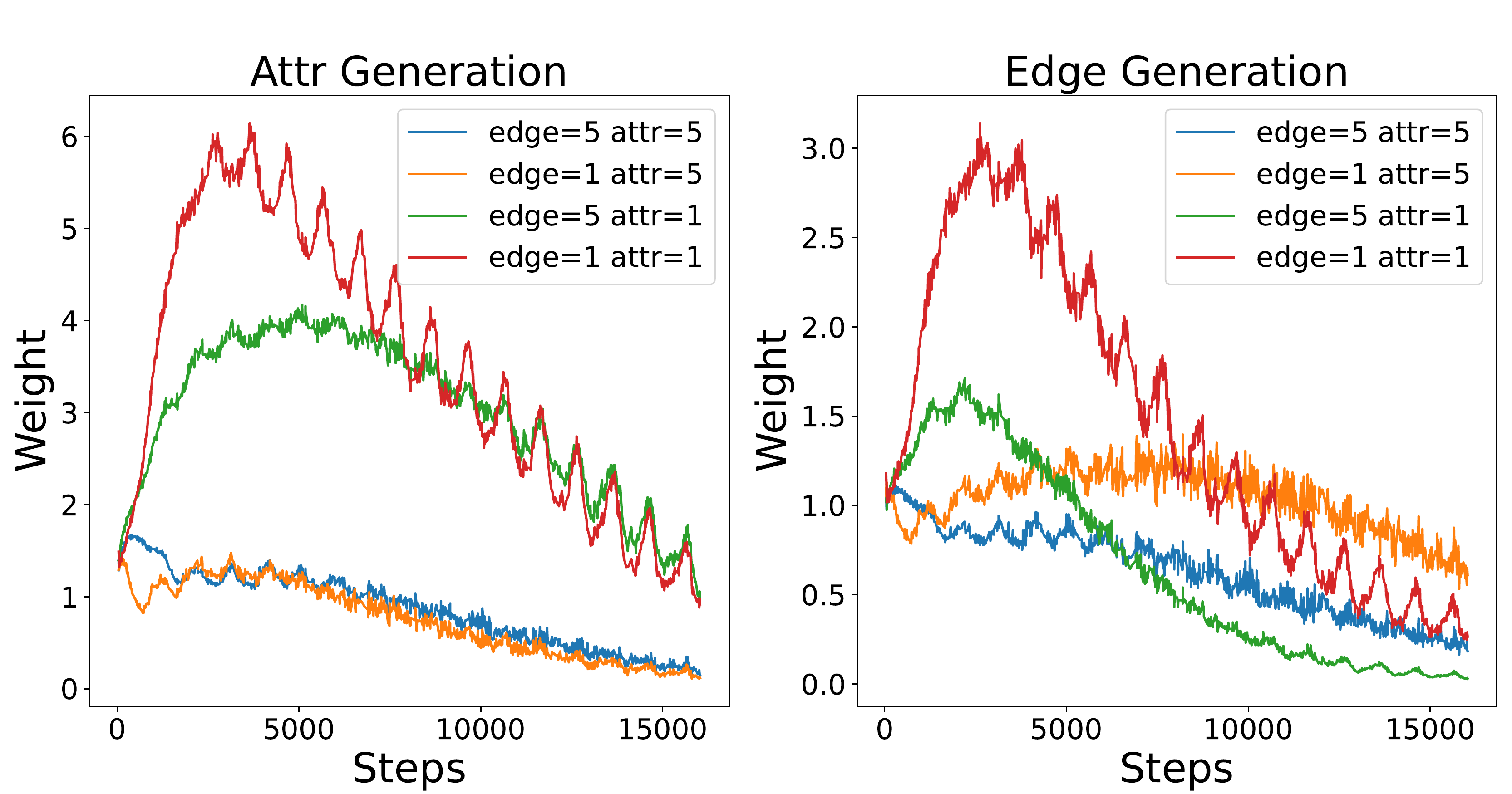}
  \caption{Auxiliary task weight curves for AUX-TS on four groups of loss scales (OAG\_CS, P\&F).}
  \label{fig:rescale}
\end{figure}

\textbf{AUX-TS performs well regardless of the loss scale.} 
To further prove advantages of AUX-TS over SELAR which only uses loss in the weighting model without gradient similarity, we adjust loss scales of the two auxiliary tasks. Table~\ref{tab:ablation} shows that SELAR is very sensitive to the loss scale and performs poorly on readjusted losses, while AUX-TS consistently performs well in various loss scale settings.

As shown in Figure~\ref{fig:rescale}, by comparing the weight curves under different loss scales, it is proved that AUX-TS could adaptively balance weights of different tasks under different loss scales. It assigns lower weights to tasks with high loss scales, which is different from SELAR. We observe a certain consistency between weight and loss in SELAR. SELAR tends to give high weights to hard examples with a large loss more than easy ones. However, the loss which represents the difficulty of a task is insufficient to characterize the importance of the task. The reason why AUX-TS could outperform SELAR significantly is that it not only consider the difficulty of a task but also consider the consistency with the target task.

\section{Conclusion}
In this work, we propose a new generic transfer learning paradigm on GNNs which would adaptively add multiple types of auxiliary tasks in the fine-tuning stage to continually incorporate various priors . It can improve the effectiveness of knowledge transfer and narrow the gap between pre-training and fine-tuning. Furthermore, we design an adaptive auxiliary loss weighting model to learn the weights of auxiliary tasks by quantifying the consistency between auxiliary tasks and the target task. The model is learnable through meta-learning. Comprehensive experiments on various downstream tasks and transfer learning schemes prove the generality and superiority of our methods compared with other methods. The effectiveness of our approach is demonstrated on GNNs but not limited to GNNs scenarios. It's a generic transfer learning paradigm that could be flexibly extended to various transfer learning scenarios such as in NLP or computer vision (CV).

\bibliographystyle{ACM-Reference-Format}
\bibliography{kdd21.bib}

\newpage
\appendix
\section{Implementation Details}
The weighting model is a 2-layer MLP network, the first layer uses ReLU as the activation function, and the last layer uses softplus. We set the hidden dimensions to 1000.

For OAG\_CS and Reddit datasets, we 
run experiments with three seeds and evaluate 40 batches for each run to get more stable results.

For hyper-parameters used in OAG\_CS and Reddit, we set the batch size to 256, the learning rate for GNNs to 1e-3/5e-4 respectively for pre-training and fine-tuning, and set the learning rate for the weighting model to 1e-5. We tried sigmoid and softplus as the activation function for the last layer of the weighting model. In most cases, softplus can achieve better performance. We do not use cross validation for experiments on OAG\_CS and Reddit, but use 3-fold cross validation on Last-FM and Book-Crossing datasets.

\section{Auxiliary Tasks Details}
We use five meta-path prediction tasks as auxiliary tasks for Last-FM and Book-Crossing datasets, and use the meta-paths label data published by SELAR. According to SELAR, these auxiliary tasks are listed in ~\ref{tab:MP}.
\begin{table}[h]
\centering
\caption{Auxiliary tasks on \textbf{Last-FM} and \textbf{Book-Crossing} datasets.}
\label{tab:MP}
\begin{tabular}{l|l}
\toprule
Meta-paths (Last-FM) & Meta-paths (Book-Crossing) \\
\midrule
user-item-actor-item & user-item-literary.series-item-user\\
user-item-appearing.in.film-item & item-genre-item\\
user-item-instruments-item & user-item-user-item\\
user-item-user-item & user-item-user\\
user-item-artist.origin-item & item-user-item\\
\bottomrule
\end{tabular}
\end{table}
\end{document}